\documentclass[10pt,twocolumn,letterpaper]{article}

\usepackage{wacv}
\usepackage{times}
\usepackage{epsfig}
\usepackage{graphicx}
\usepackage{amsmath}
\usepackage{amssymb}
\usepackage{subfig}
\usepackage{multirow}
\usepackage{algpseudocode}
\usepackage{balance}
\usepackage{bm}
\usepackage[hidelinks=true]{hyperref}
\hypersetup{
  colorlinks=false,
  pdfborder={0 0 0}
}

\wacvfinalcopy %

\def\wacvPaperID{191} %

\ifwacvfinal\fi
\begin{document}

\title{Weakly-Supervised Spatial Context Networks}

\author{Zuxuan Wu \\
University of Maryland\\
{\tt\small zxwu@cs.umd.edu}
\and
Larry S. Davis \\
University of Maryland\\
{\tt\small lsd@umiacs.umd.edu}
\and 
Leonid Sigal \\
University of British Columbia \\
{\tt \small lsigal@cs.ubc.ca}
}

\maketitle
\ifwacvfinal\thispagestyle{empty}\fi

\begin{abstract}
  We explore the power of spatial context as a self-supervisory signal for learning
  visual representations. 
  In particular, we propose spatial context networks that learn to predict a 
  representation of one image patch from another image patch, within the same 
  image, conditioned on their real-valued relative spatial offset. 
  Unlike auto-encoders, that aim to encode and reconstruct original image patches, 
  our network aims to encode and reconstruct {\em intermediate} representations 
  of the {\em spatially offset} patches.    
  As such, the network learns a spatially conditioned contextual representation.
  By testing performance with various patch selection mechanisms 
  we show that focusing on object-centric patches is important, and   
  that using object proposal as a patch selection
  mechanism leads to the highest improvement in performance.
  Further, unlike auto-encoders, context encoders~\cite{Pathak2016}, or other forms of %
  unsupervised feature learning, we illustrate that contextual supervision (with pre-trained 
  model initialization) can improve on existing pre-trained model performance. 
  We build our spatial context networks on top of standard VGG\_19 and CNN\_M 
  architectures and, among other things, show that we can achieve improvements (with no additional 
  explicit supervision) over the original ImageNet pre-trained VGG\_19 and CNN\_M models in object 
  categorization and detection on VOC2007. %

\end{abstract}

\vspace{-0.2in}

\section{Introduction}
Recent successful advances in object categorization, detection and segmentation
have been fueled by high capacity deep learning models (\eg, CNNs) learned from
massive labeled corpora of data (\eg, ImageNet \cite{Russakovsky2015}, COCO \cite{Lin2014}).
However, the large-scale human supervision that makes these methods effective at the 
same time, limits their use; especially for fine-grained object-level tasks such
as detection or segmentation, where annotation efforts become costly and
unwieldily at scale. 
One popular solution %
is to use a pre-trained model 
(\eg, VGG\_19 trained on ImageNet) %
for other, potentially unrelated, image tasks. 
Such pre-trained models produce effective and highly generic feature representations 
\cite{Donahue2014,Razavian2014}. However, it has also been shown 
that fine-tuning with task-specific labeled samples is often necessary %
\cite{girshick2014rcnn}.  

Unsupervised learning is one way to potentially address some of these challenges. 
Unfortunately, despite significant research efforts unsupervised models 
such as auto-encoders \cite{Hinton2006,Vincent2008} and, 
more recently, context encoders \cite{Pathak2016} have not produced representations that 
can rival pre-trained models (let alone beat them). 
Among the biggest challenges is how to encourage 
a representation that captures semantic-level (\eg, object-level) information 
without having access to explicit annotations for object extent or class labels. 

\begin{figure}[t!]
\centering
\includegraphics[width=0.9\columnwidth]{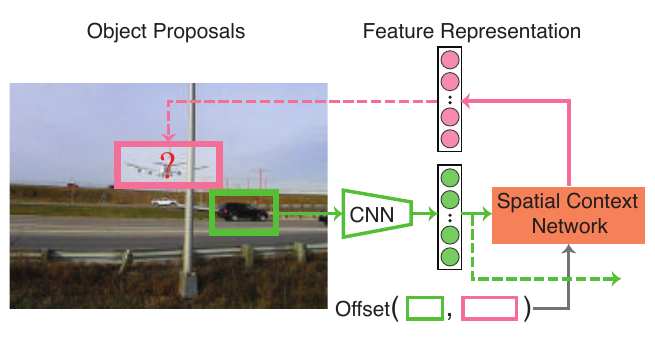}
\vspace{-0.15in}
\caption{{\bf Illustration of the proposed spatial context network.} 
A CNN used to compute feature representation 
of the green patch is fine-tuned to predict feature representation of the red patch 
using the proposed spatial context module, conditioned on their relative offset. 
Pairs of patches used to train the network are obtained from object proposal mechanisms. 
Once the network is trained, the green CNN can be used as a 
generic feature extractor for other tasks (dotted green line).}
\label{fig:framework}
\vspace{-0.25in}
\end{figure}
In the text domain, the idea of local spatial context within a sentence, proved to 
be an effective supervisory signal for learning distributed word 
vector representations (\eg, continuous bag-of-words (CBOW) \cite{Mikolov2013} 
and skip-gram models \cite{Mikolov2013}). 
The idea is conceptually simple; given a word tokenized corpus of text, 
learn a representation for a target word that allows it to predict 
representations of contextual words around it; 
or vice versa, given contextual words to predict a representation of the target word.   
Generalizing this idea to images, while appealing, is also challenging as 
it is not clear how to 1) {\em tokenize} the image (\ie, what is an elementary 
entity between which context supervision should be applied) and 
2) apply the notion of context effectively in a 2-D real-valued 
domain. 

Recent attempts to use spatial context as supervision in vision, resulted in models that used 
(regularly sampled) image patches as {\em tokens} and either learned a representation that is useful for classifying contextual relationships 
between them \cite{Doersch} or attempted to learn representations that fill in an image patch based on the larger surrounding 
pixels \cite{Pathak2016}. 
In both cases, the resulting feature representations fail to perform at the level of the pre-trained 
ImageNet models.
This could be attributed to a number of reasons: 
1) spatial context may indeed not be a good supervisory signal; 
2) generic and neighboring image patches may not be an effective {\em tokenization} scheme; and/or 
3) it may be difficult to train a model with a contextual loss from scratch.

Our motivation is similar to \cite{Doersch,Pathak2016}; 
however, we posit that image {\em tokenization} is important and 
should be done at the level of objects.
By working with patches at object scale, 
our network can focus on more object-centric features and 
potentially ignore some of the texture and color detail that are
likely less important for semantic tasks.
Further, instead of looking at immediate regions around the patch for 
context \cite{Pathak2016} and encoding the relationship between the 
contextual and target regions implicitly, we look at potentially
non-overlapping patches with longer spatial contextual dependencies 
and explicitly condition the predicted representation on the 
relative spatial offset between the two regions. 
In addition, when training our network, we make 
use of a pre-trained model to extract intermediate representations. 
Since lower levels of CNNs have been shown to be task 
independent, this allows us to learn a better representation.

Specifically, we propose a novel architecture --
Spatial Context Network (SCN) -- which is built on top of existing CNN 
networks and is designed to predict a
representation of one (object-like) image patch from another 
(object-like) image patch, conditioned on their relative spatial offset. 
As a result, the network learns a spatially conditioned contextual 
representation of image patches. 
In other words, given the same input patch and different spatial offsets 
it learns to predict different contextual representations (\eg, given a 
patch depicting a side-view of a car and a horizontal offset, 
the network may output a patch representation of another car; however, the 
same input patch with a vertical offset may result in a patch representation of a plane).  
We also make use of ImageNet pre-trained model as both
an initialization and to define intermediate representations. 
{\color{black}
Once an SCN model is trained (on pairs of patches), we can use one of its 
two streams as an image representation that can be used for a 
variety of tasks, including object categorization or localization (\eg, as part of Faster R-CNN \cite{girshick2015fast}). }
This setting allows us to definitively answer the question of whether spatial 
context can be an effective supervisory signal -- it can, improving
on the original ImageNet pre-trained models.

\vspace{0.1in}
\noindent
{\bf Contributions: }
Our main contribution is the spatial context network (SCN), which
differs from other models in that it uses two offset patches
as a form of contextual supervision. 
Further, we explore a variety of tokenization schemes for mining training patch pairs, 
and show that an object proposal mechanism is the most effective.
This observation validates the intuition that for semantic tasks, context is most 
useful at the object scale. 
Finally, we conduct extensive experiments to investigate the capacity
of the proposed SCN for capturing context information in images,
and demonstrate its ability to improve, in an unsupervised manner, on ImageNet 
pre-trained CNN models for both categorization (on VOC2007 and VOC2012) and detection 
(on VOC2007), where the bottom stream of the trained SCN is used as a generic
feature extractor (see Fig.~\ref{fig:framework} (bottom)).

\section{Related Work}

\noindent{\bf Unsupervised Learning}.
Auto-encoders \cite{hinton2007learning} are among the earliest models for unsupervised 
deep learning. They typically learn a representation by employing an encoder-decoder
architecture, which are inverses of one another; 
the encoder encodes
the image (or patch) into a compact hidden state representation and the decoder 
reconstructs it back to a full image. 
De-noising auto-encoders \cite{Vincent2008} reconstruct images (or patches)
subject to local corruptions. %
The most extreme variant of de-noising auto-encoders are the context encoders 
\cite{Pathak2016}, which aim to reconstruct a large hole (patch) given its 
surrounding spatial context.

A number of papers proposed to learn representations by converting the 
generative auto-encoder-like objectives to discriminative classification
counterparts, where CNNs %
have been shown to learn effectively. 
For example, \cite{Dosovitskiy2015} proposed an idea of surrogate 
classes that are formed by applying a variety of transformations to randomly
sampled image patches. Classification into these surrogate classes is used 
as a supervisory signal to learn image representations.
Alternatively, in \cite{Doersch}, neighboring patches are used in Siamese-like networks  
to predict the relative {\em discrete} (\eg, to the top-right, bottom-left, \etc) 
location of patches.
Related, is also \cite{Zagoruyko2015} that attempts to learn a similarity function
across patches using various deep learning architectures, including 
center-surround (similar to \cite{Pathak2016}) and forms of Siamese networks. 
Goodfellow \etal~\cite{goodfellow2014generative} proposed Generative Adversarial Networks (GAN)
that contain a generative model and discriminative model. 
Pathak \etal \cite{Pathak2016} built upon GANs to model context through inpainting missing patches.

Our model is related to auto-encoders \cite{hinton2007learning}, and particularly context encoders 
\cite{Pathak2016}, however, it is conceptually between the discriminative and 
generative forms discussed above. We have encoder and decoder components, 
but instead of decoding the hidden state all way to an image, our decoder decodes
it to an intermediate discriminatively trained representation. Further, unlike
previous methods, our decoder takes real-valued patch offsets
as input, in addition to the representation of the patch itself. 

\vspace{0.05in}
\noindent
{\bf Pre-trained Models}. 
Pre-trained CNN models have been shown to generalize to a large number of different 
tasks \cite{Donahue2014,Razavian2014}. 
However, their transferability, as was noted in \cite{Yosinski2014}, is affected
by specialization of higher layer neurons to the original task (often ImageNet categorization). 
By taking a network pre-trained on the ImageNet task and using its intermediate
representation as target for our decoder, we make use of the knowledge 
distilled in the network \cite{hinton2015distilling} while attempting to improve it using 
spatial context.  
Works like \cite{Oquab2014} and \cite{Hoffman2014} attempt to similarly re-use
lower layers \cite{Oquab2014} of the pre-trained network and fine-tune, typically,
fully-connected layers to specific tasks (\eg, object detection). 
However, such models assume some labeled data in the target domain, if not
for classes of interest \cite{Oquab2014}, then for related ones \cite{Hoffman2014}. 
In our case, we assume no supervision of this form. 
Instead, we just assume that there exists a process that can generate category 
agnostic object-like proposal patches. Our work is similar to~\cite{zhangaugmenting} that 
also attempts to improve the performance of pre-trained models. 
While they augment existing networks with reconstructive decoding pathways for image reconstruction,
our model focuses on exploiting contextual relationships in images.

\vspace{0.05in}
\noindent {\bf Weakly-supervised and Self-supervised Learning}.
Weakly-supervised and self-supervised learning attempt to achieve similar performance 
to fully supervised models with limited use of annotated labels. 
A typical setting is to, for example, use image-level annotations
to learn an object detection model \cite{Cinbis2014,Oquab2015,Shi2013,SongICML2014,SongNIPS2014,Wang2014}. 
However, such models typically rely on latent variables and appearance regularities 
present within individual object class. In addition, researchers also utilized 
motion coherence (tracked patches~\cite{wang2015unsupervised}
or ego-motion from sensors~\cite{agrawal2015learning}) in videos as supervisory signals to train networks. 
Zhang \etal~\cite{zhang2016colorful} generated a color version of a grayscale 
photo through a CNN model,
which could further serve as an auxiliary task for feature learning. Noroozi \etal learned features by solving jigsaw puzzles~\cite{noroozi2016unsupervised}.
Different from these works, we experiment with (category-independent) object proposals as a way to 
{\em tokenize} an image into more semantically meaningful parts. This can
be thought of as (perhaps) a very weak form of supervision, but unlike any 
that we are aware has been used before.  

{\color{black}
Also related is \cite{Vondrick2016}, where the model for predicting future frame representation 
in video, given the current frame representation, is learned. The premise in \cite{Vondrick2016}
is conceptually similar to ours, but there are important differences. 
Our predictions are on spatial category-independent object proposals (not frames offset in 
time \cite{Vondrick2016}). Further, our neural network architecture is parametrized by
the real-valued offset between pairs of proposals, where as temporal
offset in \cite{Vondrick2016} is not part of the model and is fixed to $1$ second. 
}

\begin{figure}[t!]
\centering
\includegraphics[width=0.77\linewidth]{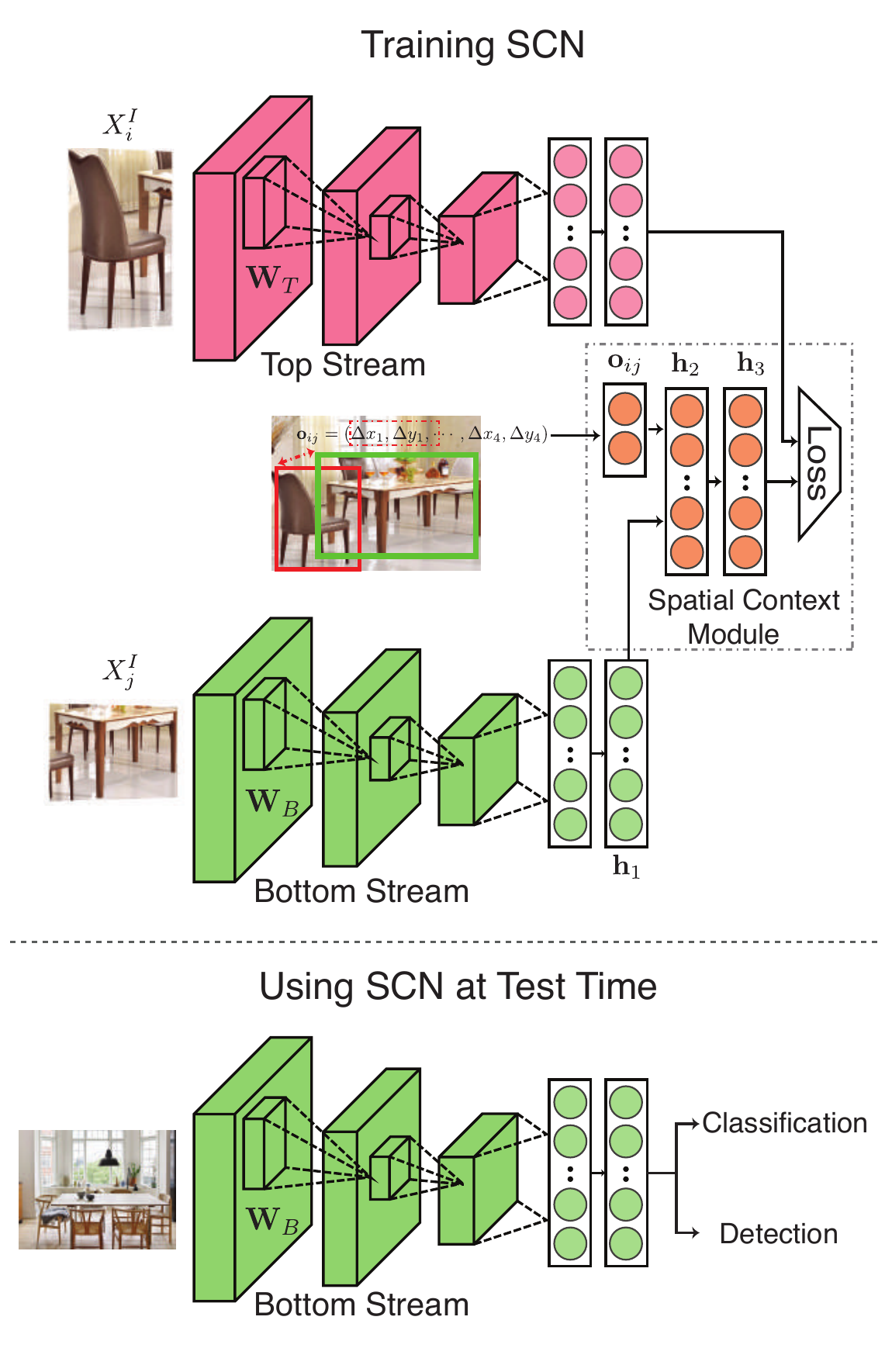}
\caption{{\bf Overview of the proposed spatial context network architecture.} See texts for complete description and discussion.}
\label{fig:framework}
\end{figure}

\section{Spatial Context Networks} %
We now introduce the proposed spatial context network (see Figure~\ref{fig:framework} (top)), which consists of a top stream and a bottom stream operating on a pair of patches cropped from the same image. The goal is to utilize their spatial layout information as contextual clues for feature representation learning. 
{\color{black} Once the spatial context network is learned, the bottom stream can be used as a feature extractor (see Figure~\ref{fig:framework} (bottom)) for a variety of image recognition tasks, specifically, object categorization and detection. }

More formally, given a patch $\mathbf{X}^I_i$ extracted from an image $I \in \mathbb{I}$, where $\mathbb{I}$ is the training set, we denote the patch bounding box $\mathbf{b}^I_i$ as an eight-tuple consisting of $(x,y)$ positions of top-left, top-right, bottom-left and bottom-right corners. 
We can then denote the training samples for the network as 3-tuples ($\mathbf{X}^I_i$, $\mathbf{X}^I_j$, ${\bf o}^I_{ij}$), where ${\bf o}^I_{ij} = \mathbf{b}^I_i - \mathbf{b}^I_j$ is the relative offset between two patches computed by subtracting locations of their respective four corners.

\vspace{0.05in}
\noindent\textbf{Top stream}. The goal of the top stream is to provide a feature representation for patch $\mathbf{X}^I_i$ that will be used as soft {\em target} for contextual prediction by the {\em learned} representation of the patch $\mathbf{X}^I_j$.
This stream consists of an ImageNet pre-trained state-of-the-art CNN such as VGG\_19, GoogleNet or ResNet (any pre-trained CNN model can be used).
More specifically, the output of the top stream is the representation from the fully-connected layer ($fc_7$) obtained by propagating patch $\mathbf{X}^I_i$ through the original pre-trained ImageNet model (here we remove the softmax layer).
More formally, let $g(\mathbf{X}^I_i; {\bf W}_T)$ denote the non-linear function approximated by the CNN model and parameterized by weights ${\bf W}_T$. 
Note that one can also utilize representation of other layers; we use $fc_7$ for simplicity and because of its superior performance in most high-level visual tasks~\cite{Razavian2014}.

\vspace{0.05in}
\noindent\textbf{Bottom stream}. The bottom stream consists of an identical CNN model to the top stream which feeds into the spatial context module. The spatial context module then accounts for spatial offset between the input pair of patches. 
{\color{black} 
The network first maps the input patch to a feature representation ${\bf h}_1 = g(\mathbf{X}^I_j;{\bf W}_B)$ and then the resulting ${\bf h}_1$ ($fc_7$ representation) is used as input for the spatial context module. We initialize the bottom stream with the ImageNet pre-trained model as well, so initially, ${\bf W}_B = {\bf W}_T$. However, while ${\bf W}_T$ remains fixed, ${\bf W}_B$ is optimized during training.  
}

\vspace{0.05in}
\noindent\textbf{Spatial Context Module}. The role of the spatial context module is to take the feature representation of the patch $\mathbf{X}^I_j$ produced by the bottom stream and, given the offset to patch $\mathbf{X}^I_i$, predict the representation of patch $\mathbf{X}^I_i$ that would be produced by the top stream. 
The spatial context module is represented by a non-linear function $f([{\bf h}_1,  {\bf o}^I_{ij}]; {\bf V})$, parameterized by weight matrix ${\bf V} = \{ \mathbf{V}_1, \mathbf{V}_{loc}, \mathbf{V}_2 \}$.

In particular, the spatial context module first takes the feature vector ${\bf h}_1$ (computed from patch $\mathbf{X}^I_j$) together with the offset vector ${\bf o}_{ij}$ between $\mathbf{X}^I_j$ and $\mathbf{X}^I_i$ to derive an encoded representation:
\begin{equation}
    \label{eq:merge}
    {\bf h}_{2} = \sigma({\bf V}_1{\bf h}_1 + {\bf V}_{loc}{\bf o}_{ij}),
\end{equation}
where ${\bf V_1}$ denotes the weights for ${\bf h}_1$; ${\bf V}_{loc}$ is the weight matrix for the input offset, and $\sigma(x) = 1/(1+e^{-x})$. (Note that we absorb the bias term in the weight matrix for convenience). Finally, ${\bf h}_2$ is mapped to ${\bf h}_3$ with a linear transformation to reconstruct the $fc_7$ feature vector computed by the top stream on the patch $\mathbf{X}^I_i$.

\vspace{0.05in}
\noindent \textbf{Loss Function}. Given the output feature representations from the aforementioned two streams, we train the network by regressing the features from the bottom stream to those from the top stream. We use a squared loss function:
\begin{equation}
    \label{eq:loss}
    \min_{{\bf V}, {\bf W}_B} \sum_{I\in\mathbb{I}; i \neq j}\left\|g(\mathbf{X}^I_i;{\bf W}_T) -  f([g(\mathbf{X}^I_j;{\bf W}_B),  {\bf o}_{ij}]; {\bf V})\right\|^2.
\end{equation}
The model is essentially an encoder-decoder framework with the bottom stream encoding the input image patch into a fixed representation and 
spatial context module decoding it to representation of another, spatially offset, patch. 
The intuition comes from the skip-gram model~\cite{mikolov2013efficient} that attempts to predict the context given a word, which has been demonstrated to be effective for a number of NLP tasks. Since objects often co-occur in images in particular relative locations, it makes intuitive sense to explore such relations as contextual supervision.

The network can be easily trained using back-propagation with stochastic gradient descent. Note that for the top stream, rather than predicting raw pixels in images, we utilize the features extracted from off-the-shelf CNN architecture as {\em ground truth}, to which the features constructed by the bottom stream regress. This is because the pre-trained CNN model contains valuable semantic information (\eg, referred to as dark knowledge \cite{hinton2015distilling}) to differentiate objects and the extracted off-the-shelf features have achieved great success on various tasks~\cite{zha2015exploiting,zhou2014learning}. 

One alternative to formulating the problem as a regression task would be to turn it into a classification problem by appending a softmax layer on top of the two streams and predicting whether a pair of features is likely given the spatial offset. 
However, this would require a large number of negative samples (\eg, a \emph{car} is not likely to be in a \emph{lake}), making training difficult. Further, our regression loss also builds on intuitions explored in \cite{hinton2015distilling}, where it is shown that soft real-valued targets are often better than discrete labels.

\vspace{0.05in}
\noindent\textbf{Implementation Details}. We adopt two off-the-shelf CNN architectures, CNN\_M and VGG\_19~\cite{simonyan2014very}, to train the spatial context network. CNN\_M is an AlexNet~\cite{krizhevsky2012imagenet} style CNN with five convolutional layers topped by three fully-connected layers (the dimension for $fc_6$ and $fc_7$ is $2,048$), but contains more convolutional filters. VGG\_19 network consists of 16 convolutional layers followed by three fully-connected layers, possessing stronger discriminative power.

The pipeline was implemented in Torch and we apply mini-batch stochastic gradient descent in training with the batch size of 64. The weights for the spatial context module are initialized randomly. We fine-tune the fully-connected layers in the bottom stream CNN model with convolutional layers fixed, unless otherwise specified. The input patches are resized to 224$\times$224. We set the initial learning rate to $1e^{-3}$, which is decreased to $1e^{-4}$ after 100 epochs; we fix weight decay to $5e{-4}$ and the maximum number of epochs to 200. We will discuss patch selection in Experiements. 

{\color{black} 
\subsection{Using SCN for Classification and Detection}
Once the SCN is trained, we {\em only} use ${\bf h}_1$ from the bottom stream as {\em a feature representation} for other tasks (Figure~\ref{fig:framework} (bottom)). %
As we will show, these feature representations are better than those obtained from the original ImageNet pre-trained model for object detection and classification. %
}

\section{Experiments}
We first validate the ability of the proposed SCN to learn context information on a synthetic dataset and with the real images from VOC2012. We then evaluate the effectiveness of features extracted from the spatial context framework in classification and detection tasks, as compared with original pre-trained ImageNet features, and competing state-of-the-art feature learning methods. %

\subsection{Synthetic Dataset Experiments} 
\begin{figure}[h]
\centering
\includegraphics[width=0.7\linewidth]{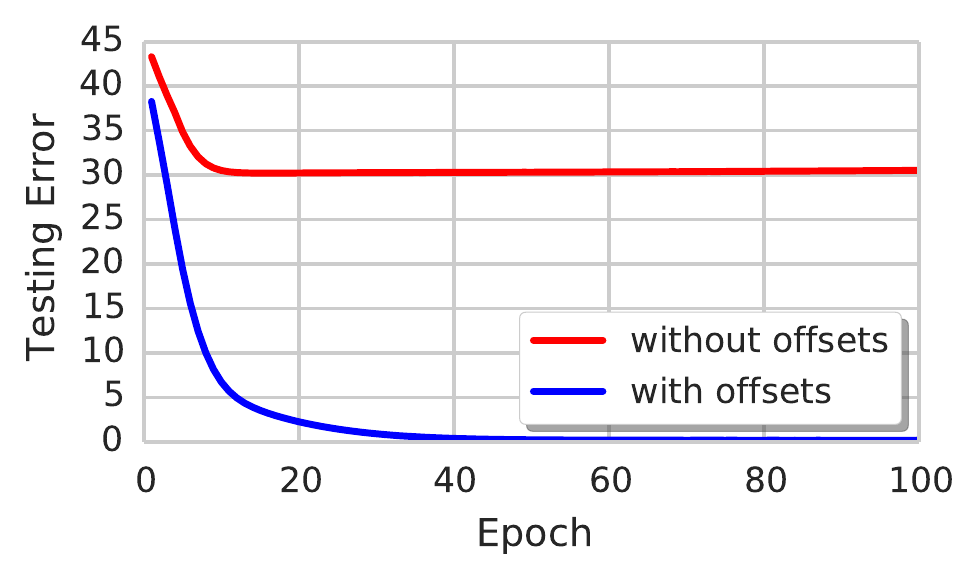}
\vspace{-0.2in}
\caption{{\bf Testing error on the synthetic dataset.} Illustrated is the testing error with and without offset vector.}
\vspace{-0.1in}
\label{fig:lossoverepoch}
\end{figure}

\begin{figure}[t!]
\centering
\includegraphics[width=1\linewidth]{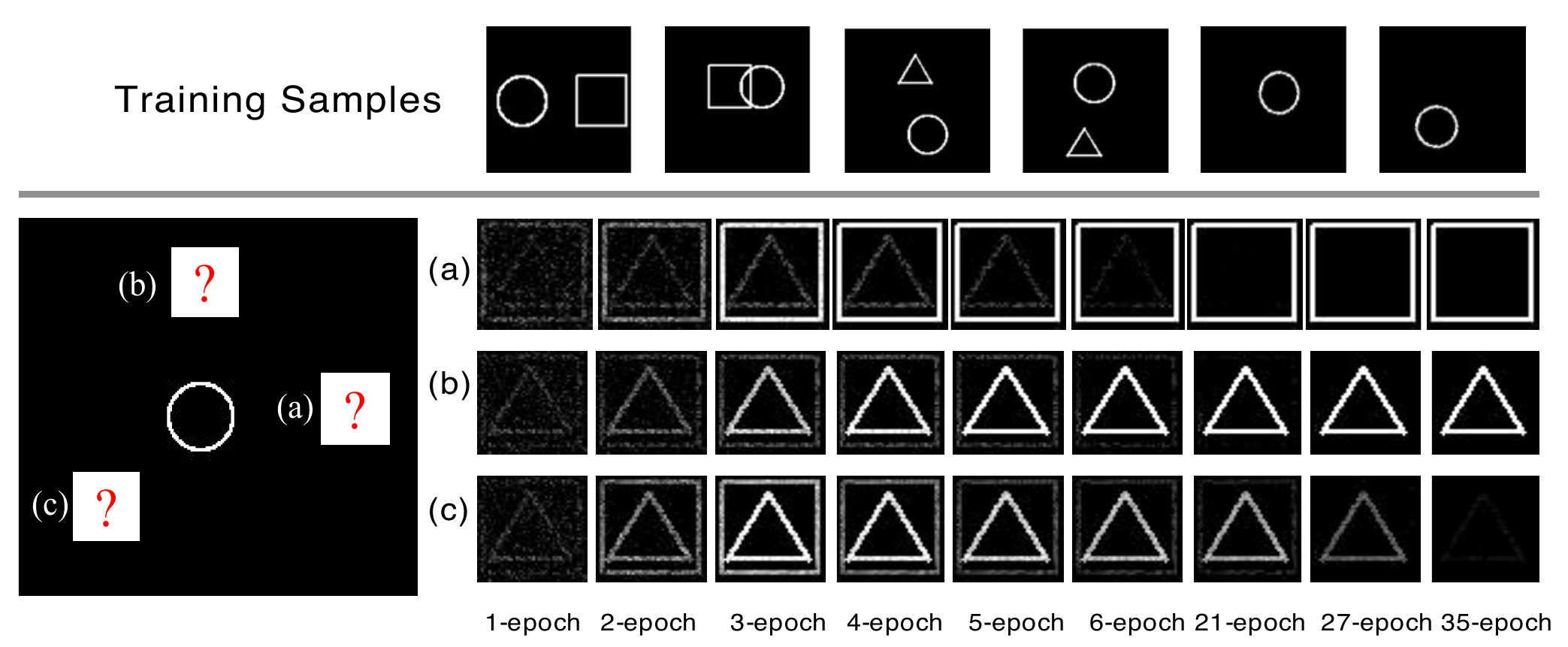}
\caption{{\bf Experiments with synthetic dataset.} Training samples are shown in top row. Bottom rows show predicted patches for the labeled regions on the left, after 1--35 epochs of training. Predicted patches are obtained by treating the circle in the middle and an appropriate spatial offset to (a), (b), or (c) as input to an SCN and visualizing the output $\mathbf{h}_3$ layer.}
\label{fig:constrcuted}
\end{figure}

We construct a synthetic dataset containing {\em circles}, {\em squares} and {\em triangles} to verify whether the proposed spatial context framework is able to learn correlations in spatial layout patterns of these objects. More specifically, we create 300 (circle, square) pairs where circles are always horizontally offset (see Figure~\ref{fig:constrcuted} (top)) from the squares (vertical difference is within 30 pixels); and 300 (circle, triangle) pairs where circles are vertically offset from the triangles (horizontal difference is within 30 pixels); as well as 200 (circles, black image) pairs where the offset vector is randomly sampled. We randomly split the dataset into 600 training and 200 testing pairs. We assume perfect proposals and crop patches tightly around the objects (circles, squares and triangles). Here, we adopt the CNN\_M model only. %

The testing error loss (mean squared error) on this dataset is visualized in Figure~\ref{fig:lossoverepoch}. As we can see from the figure, the testing error of the spatial context network steadily decreases for the first 20 epochs and nearly reaches zero after 25 epochs. To investigate the role offset vectors play in the learning process, we remove the offset vector from the input and retrain the network. The loss of this network stabilizes to 30 after 10 epochs; this is significantly higher than the error of the spatial context network. Figure~\ref{fig:lossoverepoch} confirms that the proposed spatial context network can make effective use of the spatial context information between objects.

To gain further insights into the learning process, we replace the {\em target} features of the top stream with raw ground truth image patches. After each epoch, given an input bottom stream object patch (depicting circle) and an offset vector from the testing set, we adopt the output of the last layer ${\bf h}_3$ in the SCN to reconstruct images for the top stream (See Supple. for details). The results are visualized in Figure~\ref{fig:constrcuted}. %

When circles are combined with either horizontal or vertical offsets, the network is able to reconstruct square and triangle patches (respectively) after about five epochs of training. For the first few epochs, both triangles and squares co-occur in the constructed images, but clear square and triangle patterns emerge as the training proceeds. It took longer for the network to learn that conditioned on an off-axis offset vector and a circle patch it should produce an empty (black) patch image. This experiment validates that our spatial context network is able to learn correct spatially varying contextual representation based on (identical) input patch (circle) and varying offsets.
Without providing location offset information, the network overfits and simply generates a patch containing overlapping triangles and squares (which explains the poor convergence in Figure~\ref{fig:lossoverepoch}).

Imagining a circle is a car, a square a tree and the triangle (which is above circle) to be sky, this synthetic dataset provides a simplified version of spatial context information in real-world scenarios. The experiments indicate that the varying spatial contextual information among multiple objects can be learned by the SCN.

\subsection{Modeling Context in Real Images}
\label{sec:realimage}

We now discuss context modeling in real images and validate the capability of the network to capture such real-world contextual clues. To this end, we use the PASCAL VOC 2012~\cite{everingham2010pascal} dataset, which consists of a training set with 5,717 images and a validation set with 5,823 images, totaling 20 object categories (denoted by VOC2012-Img). We first crop objects from the original images on both subsets using the provided annotations of bounding boxes, which leads to 15,774 objects for training and 15,787 objects for testing (denoted by VOC2012-Obj\footnote{\label{ft:img}The difference between VOC2012-Obj and VOC2012-Img is that in the former the objects are cropped, where as in the latter they are not.}). Objects from the same image are further paired and are used as inputs for the spatial context network (SCN) together with their offset vector. In total, we obtain 34,378 training and 34,722 testing paired samples (VOC2012-Pairs).
\begin{figure}[h!]
\centering
\includegraphics[width=0.70\linewidth]{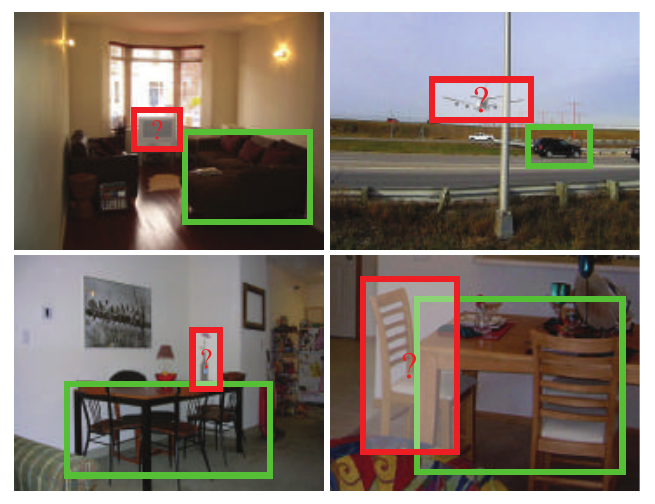}
\vspace{-0.1in}
\caption{{\bf SCN contextual classification.} Features of the top stream (red boxes) are predicted using patches from bottom stream (green boxes) and offset vector as inputs to the trained SCN. A classifier is then trained to predict the label of the red patch based on the {\em predicted} features from the training set. Performance on testing set is 56.3\% (Table~\ref{tbl:VOC2012-Bottom}).}
\label{fig:example}
\end{figure}

{ \color{black} We first train the spatial context network using paired images. Given the trained network, we compute the outputs of the last layer from the spatial context module (\ie, ${\bf h}_3$) as the synthesized feature representations for a single patch in the top stream (on both training and test set).}
Then we train a linear classifier with the extracted features using all training patches in the top stream (See Fig.~\ref{fig:example} and see Supple. for details). To establish a baseline, for all patches in the top stream, we compute the raw $fc_7$ features from the original VGG\_19 network and similarly train a linear SVM classifier. The results are summarized in Table~\ref{tbl:VOC2012-Bottom}.

It is surprising to see that the predicted features achieve a 56.3\% accuracy {\color{black} in object classification} given the fact that these features are {\em predicted} from nearby objects within the same image (from the bottom stream) using the trained spatial context network (SCN). In other words we are able to recognize objects at 56.3\% accuracy without ever seeing the real image features contained in the corresponding image patches; the recognition is done purely based on the {\em contextual} predictions of those features from other patches (note that 92.6\% of patches do not or minimally overlap ($<$ 0.2 IoU)). This indicates {\em very strong} contextual information that our network was able to learn. 
\begin{table}[t!]
\begin{center}
\small
\begin{tabular}{c|c}
\hline
features                            & VOC2012-Pairs (\%) \\ \hline   \hline 

VGG\_19 $fc_7$        & 78.3      \\ \hline         
SCN predicted (${\bf h}_3$)  features          &  56.3    \\ \hline 
VGG\_19 $fc_7$ + SCN predicted            & 79.5    \\ \hline                  
\end{tabular}
\vspace{-0.1in}
\caption{{\bf Performance comparisons of classification.} Different feature representations for the {\em top} patch classification are compared. SCN predicted features are obtained by regressing {\em top} stream features from the contextual {\em bottom} stream patch.} %
\label{tbl:VOC2012-Bottom} 
\end{center}
\vspace{-0.35in}
\end{table}

To eliminate the possibility that accuracy comes from images containing multiple instances of the same object, we analyzed the dataset and found only 45\% of training and 42\% of testing image patch pairs correspond to the same objects. Further, using pairs that do not contain same objects produces an accuracy of 52.8\%, and 63.2\% with pairs only from the same objects.

To investigate whether the synthesized features ${\bf h}_3$ contain contextual information that might be complementary to the original $fc_7$ features, we perform feature fusion by concatenating the two representations into a 8,192-D vector and training a linear SVM for classification. We observe 1.2\% performance gain compared with raw VGG $fc_7$ features, confirming context is beneficial.%

\subsection{Feature Learning with SCN for Classification}

In the last two sections, to verify the effectiveness of spatial contextual learning, we assumed 
knowledge of object bounding boxes (but, importantly, not their categorical identity); in other words, 
we assumed existence of a perfect object proposal mechanism; this is clearly unrealistic.
In this section, we explore the importance/significance of the quality of the object proposal mechanism
on the performance of features learned using SCN.  
{\color{black} We do so in the context of classification, where once SCN is trained, we 
use SVM on top of generic SCN features (see Figure \ref{fig:framework} (bottom)). }  

\begin{table}[h!]
\begin{center}
\small
\begin{tabular}{c|c|c|c}
\hline
          &   features-$fc_7$           &  VOC2012-Obj      & VOC2012-Img \\ \hline   \hline 
\multirow{5}{*}{\rotatebox[origin=c]{90}{CNN\_M}} & Original  & 75.3        &  68.5      \\  \cline{2-4}
& SCN-BBox    & 78.7        &  70.8      \\  \cline{2-4}
& SCN-YOLO    & 79.2        &  70.7      \\  \cline{2-4}
& SCN-EdgeBox   & \textbf{79.9}       &  \textbf{72.8}       \\  \cline{2-4} 
& SCN-Random    & 78.8        &  70.0    \\  \cline{2-4} \hline \hline
\multirow{5}{*}{\rotatebox[origin=c]{90}{VGG\_19}} & Original   & 81.4        &  78.1      \\ \cline{2-4}

& SCN-BBox            & 82.6            & 78.8   \\ \cline{2-4}
& SCN-YOLO  & 83.0                    & 79.0  \\ \cline{2-4}
& SCN-EdgeBox  &   \textbf{83.6}                & \textbf{79.5} \\ \cline{2-4}   
& SCN-Random    &    83.2          &  79.2     \\ \hline 
\end{tabular}
\vspace{-0.1in}
\caption{{\bf Performance with various object proposals.} Comparison of classification with features obtained using SCN trained with different patch selection mechanisms is illustrated on VOC2012-Obj and VOC2012-Img, using two CNN architectures.}
\label{tbl:bbox} 
\end{center}
\vspace{-0.3in}
\end{table}

We use ground truth bounding boxes, provided by the dataset, as a baseline ({\it SCN-BBox}). In addition, we test the following object proposal methods:
\begin{itemize}
\item[-] {\bf Random Patches} ({\it SCN-Random}): We randomly crop 5 patches of size of 64 $\times$ 64 in each image (consistent with \cite{Pathak2016}) to generate 10 patch pairs per image. In total, we collect 28K cropped patches and 57K pairs.\footnote{Note that in the pairing process one could simply swap the inputs of the top and bottom stream to double the number of pairs for the network, however, empirically, we found it not to be helpful.}
\item[-] {\bf Edge Box}~\cite{zitnick2014edge} ({\it SCN-EdgeBox}): EdgeBox is a generic method to generate object bounding box proposals based on edge responses. We filter out the bounding boxes with confidence lower than 0.1 and those with irregular aspect ratio, leading to 43K {\em object} patches and 160K pairs for training. 
\item[-] {\bf YOLO}~\cite{redmon2015you} ({\it SCN-YOLO}): YOLO is a recently introduced end-to-end framework trained on VOC for object detection. We use YOLO as an object proposal mechanism, by taking patches from detection regions but ignoring the detected labels. We collect 13K objects forming 17K image patch pairs.
\end{itemize}
\begin{figure*}[t]
\centering
\includegraphics[width=0.8\linewidth]{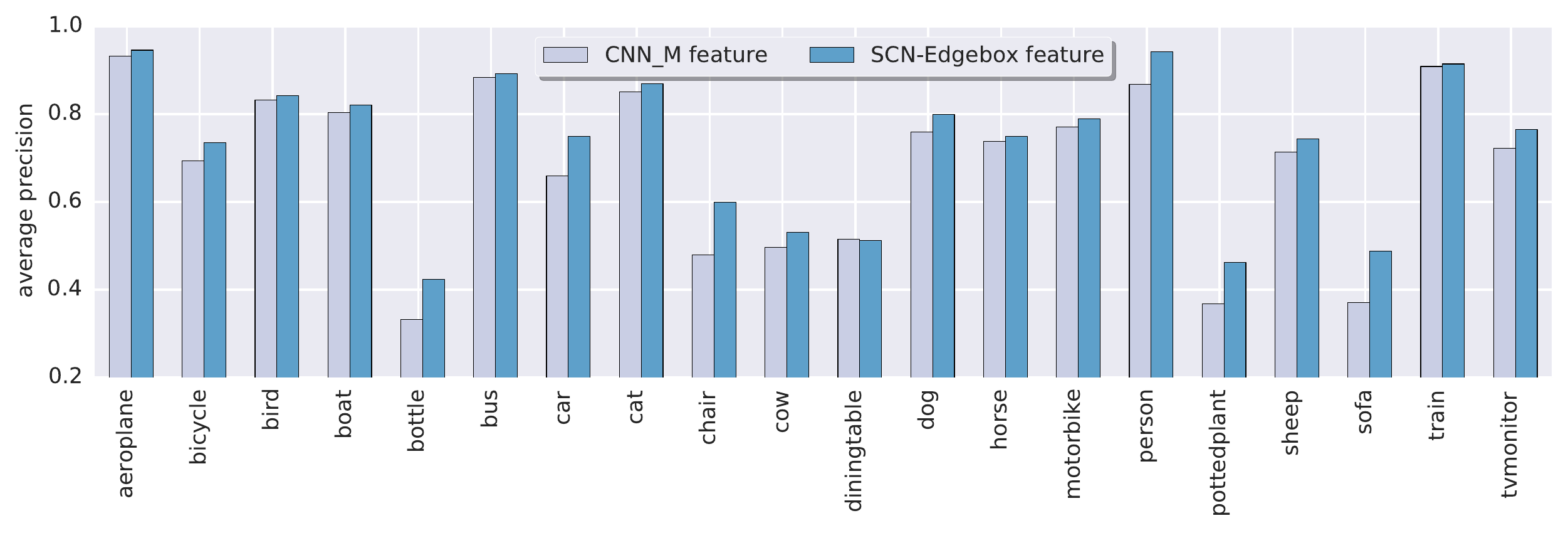}
\vspace{-0.2in}
\caption{{\bf Classification per class performance.} Reported is average precision obtained using original CNN\_M features and SCN-EdgeBox features on VOC2012-Img.}
\label{fig:perclass}
\end{figure*}
We expect the quality of object proposal methods (from least object-like to most object-like) on VOC to roughly follow the following pattern:  
\begin{center}
{\it Random}  $<$ {\it EdgeBox} $<$ {\it YOLO} $<$ ground-truth {\it BBox}. 
\end{center}

Given a trained SCN model, we utilize the bottom stream (see Fig.~\ref{fig:framework} (bottom)) to test generalization of the learned feature representations, by performing classification with linear SVMs on VOC2012-Obj and VOC2012-Img (see footnote~\ref{ft:img} for explanation) with the outputs from the first hidden layer (${\bf h}_1$, \ie, fine-tuned version of $fc_7$) in the bottom stream of SCN. The results are measured in mAP. We compare the different patch selection mechanisms discussed above and also to the original ImageNet pre-trained models.
The results are summarized in Table~\ref{tbl:bbox}. We observe that SCN-BBox and SCN-YOLO achieve better results compared with the original $fc_7$ features. It is also surprising to see that SCN-EdgeBox obtains the best performance, even higher than models trained with ground-truth bounding boxes. It is $4.6$ and $4.3$ percentage points better than
the original $fc_7$ features on VOC2012-Obj and VOC2012-Img. 

We believe that better performance of the SCN-EdgeBox stems from EdgeBox's ability to select object-like regions that go beyond the 20 object classes labeled in ground truth and detected by YOLO.
We also note that while Random patch sampling also improves the performance, with respect to the original ImageNet pre-trained network, it is doing so by a much smaller margin than EdgeBox patch sampling.

The original $fc_7$ features are trained using labels from ImageNet; our spatial context network is appealing in that it learns a better feature representation by exploiting contextual cues without any additional explicit supervision. 
Figure~\ref{fig:perclass} compares the per-class performance of SCN-EdgeBox and the original $fc_7$ features on VOC2012-Img, where we can see that SCN-EdgeBox features outperform the original $fc_7$ features for all classes. It is also interesting to see that, for small objects, such as ``bottle'' and ``potted plant'', the performance gain of SCN-EdgeBox is more significant.

\begin{table}[h!]
\begin{center}
\small
\begin{tabular}{c|c}
\hline
                          &  VOC2012-Obj    \\ \hline   

VGG\_19 $fc_7$          & 81.4        \\  \hline \hline
SCN-EdgeBox ($fc_{6}$, $fc_{7}$) &   83.6           \\ \hline
SCN-EdgeBox ($fc_{6}$, $fc_{7}$, $conv_5$) &   84.3                 \\ \hline   
SCN-EdgeBox  (all layers)&   82.5                 \\ \hline           
\end{tabular}
\vspace{-0.1in}
\caption{{\bf Exploring SCN learning strategies.} Classification performance based on features obtained using different fine-tuning strategies. See text for more details.}
\label{tbl:fc6} 
\end{center}
\end{table}

\noindent \textbf{Fine-tuning Convolutional Layers}. In addition to only fine-tuning the fully-connected layers of the bottom stream CNN model, we also explore whether joint training with VGG\_19 network could further improve the performance of the extracted features. More specifically, for the top stream we fix the weights since computing features dynamically poses challenges for network convergence. Further, this avoids {\em trivial} solutions of both streams learning, for example, to predict zero features for all patches. In addition, this makes use of transferability of lower levels of pre-trained CNN models as targets for the bottom stream decoding. The results are summarized in Table~\ref{tbl:fc6}. By back-propagating the error through deeper layers we observe a significant performance gain (2.9 percentage points) over the original features of VGG\_19 network, which confirms the fact that SCN is effective and VGG layers could be fine-tuned jointly for specific tasks in order to gain better performance using our formulation. When fine-tuning all layers in the network, the performance of SCN degrades slightly to 82.5\%.
\begin{table*}[h!]
\begin{center}
\small
\begin{tabular}{c|ccccc}
\hline
                        & Initialization & Supervision & Pretraining time & Classification      &  Detection \\ \hline   \hline
Random Gaussian     & random &  N/A & $<$ 1 minute & 53.3 & 43.4 \\ \hline
Wang \etal \cite{wang2015unsupervised} & random  & motion & 1 week & 58.4 & 44.0 \\ \hline
Doersch \etal \cite{Doersch} & random & context & 4 weeks & 55.3 & 46.6 \\ \hline
Pathak \etal\cite{Pathak2016} & random & context inpainting  & 14 hours & 56.5   & 44.5  \\ \hline 
Zhang \etal\cite{zhang2016colorful} & random & color & -- & 65.6 & 46.9 \\ \hline \hline
ImageNet \cite{Pathak2016}   & random & 1000 class labels  & 3 days      &  78.2      & 56.8    \\ 
*ImageNet      & random & 1000 class labels  & 3 days      &  76.9      & 58.7    \\ 
*Doersch \etal \cite{Doersch} & 1000 class labels & context & -- & 65.4 & 50.4 \\ \hline
{\bf SCN-EdgeBox}  & 1000 class labels & context & 10 hours &  79.0 & {\bf 59.4} \\ \hline
\end{tabular}
\vspace{-0.1in}
\caption{Quantitative comparison for classification and detection on the PASCAL VOC 2007 test set. {\color{black} The baselines labeled with * are based on our experiments, rest taken from original papers.} }
\label{tbl:detection} 
\end{center}
\end{table*}

\begin{figure}[b!]
\centering
\includegraphics[width=1\linewidth]{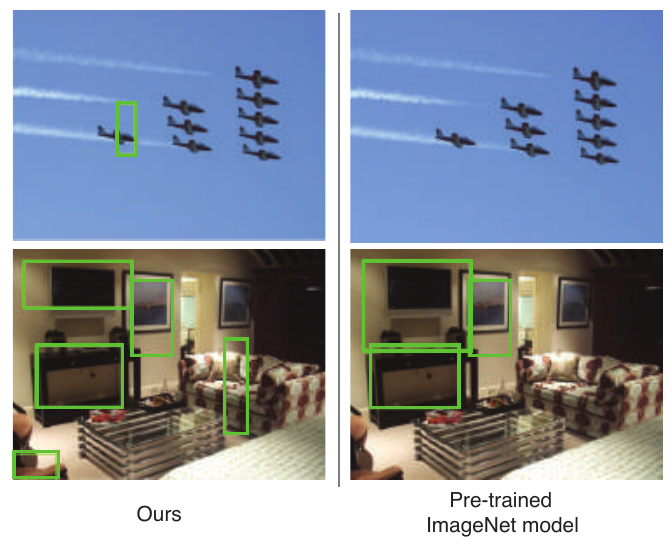}
\vspace{-0.1in}
\caption{{\bf Sample detection results.} Illustrated are results obtained using SCN-EdgeBox model and the original pre-trained ImageNet model, respectively, on VOC2007.}
\label{fig:detection}
\end{figure}

\subsection{Feature Learning with SCN for Detection}
We also explore the applicability of SCN features for object detection tasks to verify generic feature effectiveness. To make fair comparisons with prior work, 
{\color{black} we adopt the experimental setting of \cite{Pathak2016} and}
fine-tune the SCN-EdgeBox model (based on CNN\_M architecture) on Pascal VOC2007, which is then applied in the \textit{Fast R-CNN}~\cite{girshick2015fast} framework. More precisely, we replace the ImageNet pre-trained CNN\_M model with the fine-tuned bottom stream in SCN (See Figure~\ref{fig:framework} (bottom)). The weights for final classification and bounding box regression layers are initialized from scratch. Following the training and testing protocol defined in~\cite{girshick2015fast}, we finetune layers conv2 and up and report detector performance in mAP. 

The results and comparisons with existing state-of-the-art methods are summarized in Table~\ref{tbl:detection}. SCN-EdgeBox model improves on the original ImageNet pre-trained model by 0.7 percentage points. 
Further, compared with alternative unsupervised learning methods, our approach achieves significantly better performance.
We also significantly outperform other feature training methods on classification (including our fine-tuned ImageNet model)
{\color{black} and Doersch \etal \cite{Doersch} model initialized with ImageNet.}

Figure~\ref{fig:detection} visualizes some sample images where SCN-EdgeBox outperforms the pre-trained ImageNet model. Our model is better at detecting relatively small objects (\eg, airplane in the first row and chair in the second row).

\section{Conclusion}
In this paper, we presented a novel spatial context network built on top of existing CNN architectures. The SCN network exploits implicit contextual layout cues in images as a supervisory signal. More specifically, the network is trained to predict the intermediate representation of one (object-like) image patch from another (object-like) image patch, within the same image, conditioned on their relative spatial offset. Consequently, the network learns a spatially conditioned contextual representation of image patches. Extensive experiments are conducted to validate the effectiveness of the proposed spatial context network in modeling context information in images. We show that the proposed spatial context network can achieve improvements (with no additional explicit supervision) over the original ImageNet pre-trained models in object categorization on VOC2007 / VOC2012 and detection on VOC2007.

\vspace{0.05in}
{\noindent\textbf{Acknowledgment}. ZW and LSD are supported by ONR under Grant N000141612713: Visual Common Sense Reasoning for Multi-agent Activity Prediction and Recognition. LS is in part supported by NSERC Discovery grant. 
}

{\small
\bibliographystyle{ieee}
\bibliography{reference}
}

\end{document}